\documentclass[journal]{IEEEtran}
\usepackage{times}
\usepackage{amssymb}
\usepackage{siunitx}
\usepackage{booktabs}
\usepackage{array}
\usepackage{multirow}
\usepackage{placeins} % used to force float placement
\usepackage[noblocks]{authblk}
\usepackage[utf8]{inputenc}
\usepackage[T1]{fontenc}

\ifCLASSINFOpdf
\else
\fi
\usepackage{epsfig}
\usepackage{graphicx}

\usepackage{amsmath}
\usepackage{array}
\begin{document}
\title{Learning Human Identity from Motion Patterns}

\author{Natalia Neverova,
        Christian Wolf, Griffin Lacey, Lex Fridman, Deepak Chandra, \\\vspace*{-7pt} Brandon Barbello, Graham Taylor
\thanks{Manuscript created: \today.}%
\thanks{N. Neverova and C. Wolf are with INSA-Lyon, LIRIS, UMR5205, F-69621, Universit\'e de Lyon, CNRS, France. E-mail: firstname.surname@liris.cnrs.fr}%
\thanks{G. Lacey and G. Taylor are with the School of Engineering, University of Guelph, Canada. E-mail: laceyg@uoguelph.ca, gwtaylor@uoguelph.ca}%
\thanks{L. Fridman is with MIT, Boston, USA. Email: fridman@mit.edu}%
\thanks{D. Chandra and B. Barbello are with Google, Mountain View, USA. Email: dchandra@google.com, bbarbello@google.com.}%
\thanks{This work was done while N. Neverova, G. Lacey and L. Fridman were at Google, Mountain View, USA.}}


\markboth{}{}%
%{Neverova \MakeLowercase{\textit{et al.}}: Learning Human Identity from Motion Patterns}



\maketitle

\begin{abstract}
We present a large-scale study exploring the capability of temporal deep neural networks to interpret natural human kinematics and introduce the first method for active biometric authentication with mobile inertial sensors. At Google, we have created a first-of-its-kind dataset of human movements, passively collected by 1500 volunteers using their smartphones daily over several months. We (1) compare several neural architectures for efficient learning of temporal multi-modal data representations, (2) propose an optimized shift-invariant dense convolutional mechanism (DCWRNN), and (3) incorporate the discriminatively-trained dynamic features in a probabilistic generative framework taking into account temporal characteristics. Our results demonstrate that human kinematics convey important information about user identity and can serve as a valuable component of multi-modal authentication systems. Finally, we demonstrate that the proposed model can be successfully applied also in a visual context.
\end{abstract}


% \fi
\IEEEpeerreviewmaketitle

%===========================================================
\vspace{-2mm}
\section{Introduction}
\label{sec:introduction}

For the billions of smartphone users worldwide, remembering dozens of passwords for all services we need to use and spending precious seconds on entering pins or drawing sophisticated swipe patterns on touchscreens becomes a source of frustration. In recent years, researchers in different fields have been working on creating fast and secure authentication alternatives that would make it possible to remove this burden from the user \cite{sitova2015,bo2013}.
%Traditional methods, involving strong biometrics, face recognition or speaker verification, have seen a long history of prior work \cite{mccool2012, khoury2014}. 
%
%Image and speech specific descriptors have been designed and polished for decades, while an exponentially growing deep learning community has made tremendous progress in learning visual and audio representations over the last years.  On the other hand, research in deep learning has been partially driven by the demand in online applications and immediate availability of practically infinite amounts of training data.  %At the same time, there exist numerous areas where ``analytical'' descriptors have not matured, and \textit{learning} data representations is the main hope to leap beyond the current level of understanding of underlying processes. %In this sense, deep learning opens up new frontiers for learning patterns from the data which is not completely understood.  In this work, we raise the question of how to understand and interpret natural human kinematics at scale, using learned representations of data extracted from inertial physical sensors (see Fig.~\ref{fig:teaser}).  Can a smartphone, held in your hand, recognize you by your motion patterns, whether or not you are interacting with the device?

Historically, biometrics research has been hindered by the difficulty of collecting data, both from a practical and legal perspective. Previous studies have been limited to tightly constrained lab-scale data collection, poorly representing real world scenarios: not only due to the limited amount and variety of data, but also due to essential \textit{self consciousness} of participants performing the tasks.  In response, we created an unprecedented dataset of \textit{natural} prehensile movements (i.e.~those in which an object is seized and held, partly or wholly, by the hand \cite{napier1956}) collected by 1,500 volunteers over several months of daily use (Fig.~1).  %The phone's internal 3-axes accelerometer and gyroscope sensors capture continuous streams of data corresponding to 6 degrees of freedom (see Fig.~\ref{fig:teaser}).

\begin{figure}[!t] \centering 
\includegraphics[width=\linewidth]{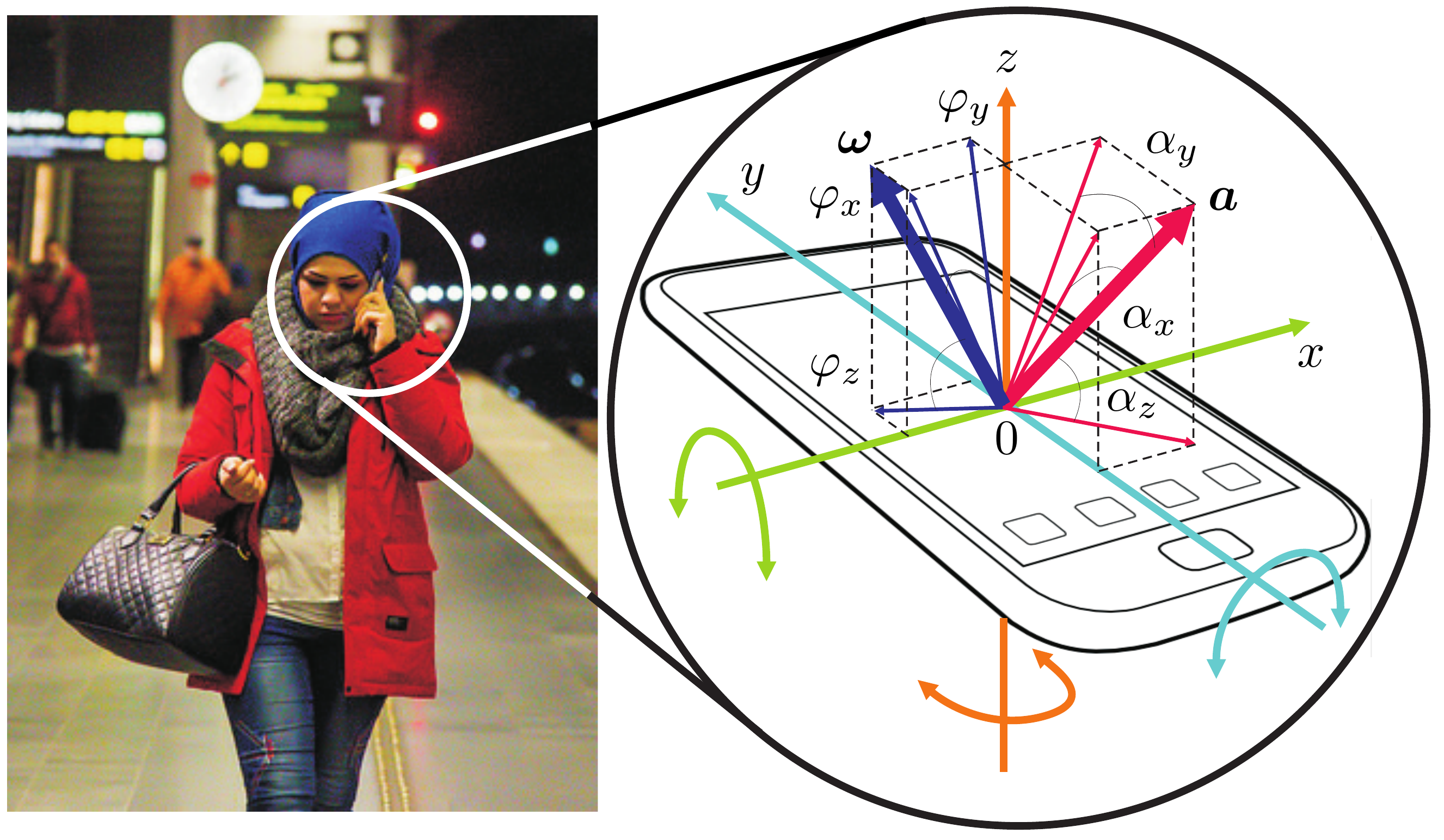}
\caption{The accelerometer captures linear acceleration, the gyroscope provides angular velocity (photo taken from \cite{photophone}).} %Right: overview of the proposed method, where the green arrow shows discriminative learning of features representations, the blue ones correspond to the optimization of a generative model, and the orange paths indicate model adaptation to a new device and test.}
\label{fig:teaser}%\vspace*{-10pt}
\end{figure}

%You can say that a goal was to create a "deep model" which is discriminatively pretrained on a large set of training devices. The next was full embedded processing and that retraining for each device was out of the question. For this reason, a generative model has been adopted, where a background model is learned « offline » (before deployment on the phones), and a device model is learned « online » (after installation on the device).

Apart from data collection, the main challenges in developing a \emph{continuous} authentication system for smartphones are (1) efficiently learning task-relevant representations of noisy inertial data, and (2) incorporating them into a biometrics setting, characterized by limited resources. Limitations include low computational power for model adaptation to a new user and for real-time inference, as well as the absence (or very limited amount) of ``negative'' samples.% during test time. % This led us to a generative model built on discriminatively-trained features extracted by deep neural networks.

In response to the above challenges, we propose a non-cooperative and non-intrusive method for on-device authentication based on two key components: temporal feature extraction by deep neural networks, and classification via a probabilistic generative model. We assess several popular deep architectures including one-dimensional convolutional nets and recurrent neural networks for feature extraction. However, apart from the application itself, the main contribution of this work is in developing a new shift-invariant temporal model which fixes a deficiency of the recently proposed Clockwork recurrent neural networks \cite{koutnik2014} yet retains their ability to explicitly model multiple temporal scales.
\vspace{-2mm}
\section{Related work}
\label{sec:relatedwork}

Exploiting wearable or mobile inertial sensors for authentication, action recognition or estimating parameters of a particular activity has been explored in different contexts. Gait analysis has attracted significant attention from the biometrics community as a non-contact, non-obtrusive authentication method resistant to spoofing attacks. A detailed overview and benchmarking of existing state-of-the-art is provided in \cite{nickel2011}. Derawi et al. \cite{derawi2010}, for example, used a smartphone attached to the human body to extract information about walking cycles, achieving $20.1\%$ equal error rate. 
%Tessendorf analysed parameters of periodic motion provided by inertial sensors to evaluate and improve rowing skills \cite{tessendorf2011}.

There exist a number of works which explore the problem of activity and gesture recognition with motion sensors, including methods based on deep learning. In \cite{figo2010} and \cite{bulling2014},  exhaustive overviews of preprocessing techniques and manual feature extraction from accelerometer data for activity recognition are given.
Perhaps most relevant to this study is \cite{plotz2011}, the first to report the effectiveness of RBM-based feature learning from accelerometer data, and \cite{bhattacharya2014}, which proposed a data-adaptive sparse coding framework.
Convolutional networks have been explored in the context of gesture and activity recognition \cite{duffner2014,zeng2014}. Lefebvre et al.~\cite{lefebvre2013} applied a bidirectional LSTM network to a problem of 14-class gesture classification, while Berlemont et al.~\cite{berlemont2015} proposed a fully-connected Siamese network for the same task.

We believe that multi-modal frameworks are more likely to provide meaningful security guarantees. A combination of face recognition and speech 
%\cite{mccool2012, khoury2014}, 
\cite{khoury2014}, 
and of gait and voice \cite{Vildjiounaite2006} have been proposed in
this context. Deep learning techniques, which achieved early success
modeling sequential data such as motion capture~\cite{Taylor2011b} and
video \cite{karpathy2014large} have shown promise in multi-modal
feature learning \cite{Ngiam2011,Srivastava2013,Kahou2013,moddrop}.

\vspace{-2mm}
\section{A generative biometric framework}
\label{sec:background}

%\color{green} You can say that a goal was to create a "deep model" which is discriminatively trained for some (obvious) reasons like better classification performance bla bla bla. However, one of the goals was full embedded processing and that retraining for each device was out of the question. For this reason, a generative model has been adopted, where a background model is learned "offline" (before deployment on the phones), and a device model is learned "online" (after installation on the device). Eventually with general equations on probability distributions. Kind of, what distribution do we want to estimate from what data.  %\color{black}

%\begin{figure*}[!htb]
%\centering
%\setlength{\unitlength}{0.1\textwidth}
%\begin{picture}(10,9)
%\includegraphics[height=0.32\linewidth]{images/architecture_single.png}\hfill
%\parbox{0.4\linewidth}{\centering
%\hspace*{-20pt}
%\vspace{-10mm} % HACK
%\includegraphics[height=0.33\linewidth]{architecture_conv.eps}\hspace{30pt}
%(a)}\hspace*{25pt}
%\parbox{0.4\linewidth}{\centering
%\includegraphics[height=0.33\linewidth]{architecture_dynamic.eps}\\
%(b)}
%\caption{Learning data representations: (a) static convnet 
%directly operating on sequences, aggregating temporal statistics by
%temporal pooling; (b) explicitly modeling temporal transitions by
%with recurrent connections.}

%\label{fig:architectures}%\vspace*{-10pt}
%\end{figure*}

Our goal is to separate a user from an impostor based on a time series of inertial measurements (Fig.~\ref{fig:teaser}). Our method is based on two components: a feature extraction pipeline which associates each user's motion sequence with a collection of discriminative features, and a biometric model, which accepts those features as inputs and performs verification. While the feature extraction component is the most interesting and novel aspect of our technique, we delay its discussion to Section~\ref{sec:featurelearning}. We begin by discussing the data format and the biometric model. 
%the simpler aspects of the model: data extraction from mobile devices and the biometric model.

\vspace{-1mm}
\subsection{Movement data}
\vspace{-1mm}
% Before giving more detail on the feature learning methods we divert briefly to discuss the formation of input vectors. 

Each reading (frame) in a synchronized raw
input stream of accelerometer and gyroscope data has the form
$\{a_x,a_y,a_z,\omega_x,\omega_y,\omega_z\}\in \mathbb{R}^6$, where
$a$ represents linear acceleration, $\omega$ angular velocity and
$x, y, z$ denote projections on corresponding axes, aligned with the
phone. There are two important steps we take prior to feature extraction.

\noindent \textbf{Obfuscation-based regularization} --- it is important to differentiate between the notion of ``device'' and ``user''. In the dataset we collected (Section \ref{sec:database}), each device is assigned to a single user, thus all data is considered to be authentic. However, in real-world scenarios such as theft, authentic and imposter data may originate from the same device.

  In a recent study \cite{das2015}, it was shown that under lab conditions a particular \textit{device} could be identified by a response of its motion sensors to a given signal. This happens due to imperfection in calibration of a sensor resulting in constant offsets and scaling coefficients (gains) of the output, that can be estimated by calculating integral statistics from the data. Formally, the measured output of both the accelerometer and gyroscope can be expressed as follows \cite{das2015}:
\begin{equation} \mathbf{a} = \mathbf{b}_a + \text{diag}(\boldsymbol{\gamma}_a)\tilde{\mathbf{a}},\quad \boldsymbol{\omega} = \mathbf{b}_{\omega} + \text{diag}(\boldsymbol{\gamma}_{\omega})\tilde{\boldsymbol{\omega}},
 \end{equation} where $\tilde{\mathbf{a}}$ and $\tilde{\boldsymbol{\omega}}$ are real acceleration and angular velocity vectors, $\mathbf{b}_a$ and $\mathbf{b}_{\omega}$ are offset vectors and $\boldsymbol{\gamma}_a$ and $\boldsymbol{\gamma}_{\omega}$ represent gain errors along each coordinate axes.
%Obfuscation-based regularization

To partially obfuscate the inter-device variations and ensure decorrelation of user identity from device signature in the learned data representation, we introduce low-level additive (offset) and multiplicative (gain) noise per training example. Following \cite{das2015}, the noise vector is obtained by drawing a 12-dimensional (3 offset and 3 gain coefficients per sensor) obfuscation vector from a uniform distribution
$\boldsymbol{\mu} \sim \mathcal{U}_{12}[0.98,1.02]$.

  % For mentioned obfuscation purposes, i.e.~to prevent leakage of information about the hardware-specific differences between devices into the neural network, 
  %, where 3 is the number of coordinate axes).  This procedure also serves as a source of data augmentation and can be seen as an additional regularization technique.

%
%\begin{equation}
%\begin{bmatrix} %a_x \\ %a_y \\ %a_z
%\end{bmatrix} %=
%\begin{bmatrix} %b_{a,x} \\ %b_{a,y} \\ %b_{a,z}
%\end{bmatrix} %+
%\begin{bmatrix} %\gamma_{a,x} & 0 & 0\\ %0 & \gamma_{a,y} & 0\\ %0 & 0 & \gamma_{a,z}
%\end{bmatrix}
%\begin{bmatrix} %a_{x,0}\\ %a_{y,0} \\ %a_{z,0}
%\end{bmatrix}
%\end{equation}
%

\textbf{Data preprocessing} --- In addition, we extract a set of angles $\alpha_{\{x,y,z\}}$ and $\varphi_{\{x,y,z\}}$ describing the orientation of vectors $\mathbf{a}$ and $\boldsymbol{\omega}$ in the phone's coordinate system (see Fig.~\ref{fig:teaser}), compute their magnitudes $|\mathbf{a}|$ and $|\mathbf{\boldsymbol\omega}|$ and normalize each of the $x, y, z$ components. Finally, the normalized coordinates, angles and magnitudes are combined in a 14-dimensional vector $\mathbf{x}^{(t)}$ with $t$ indexing the frames.% (provided below using the notation of the accelerometer stream): % \gwt{If we need more space, as I suspect we will, we can eliminate these equations or move them to the appendix.}

\subsection{Biometric model}
Relying on cloud computing to authenticate a mobile user is unfeasible due to privacy and latency.
Although this technology is well established for many mobile services, our application is essentially different from others such as voice search, as it involves constant background collection of particularly sensitive user data. Streaming this information to the cloud would create an impermissible threat from a privacy perspective for users and from a legal perspective for service providers.
Therefore, authentication must be performed on the device and is constrained by available storage, memory and processing power. Furthermore, adapting to a new user should be quick, resulting in a limited amount of training data for the ``positive'' class. This data may not be completely representative of typical usage. For these reasons, a purely discriminative setting involving learning a separate model per user, or even fine-tuning a model for each new user would hardly be feasible. 

Therefore, we adapt a generative model, namely a Gaussian Mixture Model (GMM), to estimate a general data distribution in the dynamic motion feature space and create a \textit{universal background model} (UBM). The UBM is learned offline, i.e.~prior to deployment on the phones, using a large amount of pre-collected training data. For each new user we use a very small amount of enrollment samples to perform online (i.e.~on-device) adaptation of the UBM to create a \textit{client model}. The two models are then used for real time inference of trust scores allowing continuous authentication.

\textbf{Universal background model} --- let
$\mathbf{y}{=}f(\{ \mathbf{x}^{(t)}\}){\in}\mathbb{R}^N$ be a vector of features extracted from a
raw sequence of prehensile movements by one
of the deep neural networks described in Section \ref{sec:featurelearning}. Probability densities are defined over these feature vectors as a weighted sum of $M$ multi-dimensional Gaussian distributions parameterized by $\Theta{=}\{\mathbf{\mu}_i, \mathbf{\Sigma}_i, \pi_i\}$, where $\mathbf{\mu}_i$ is a mean vector, $\mathbf{\Sigma}_i$ a covariance matrix and $\pi_i$ a mixture coefficient:
\begin{gather}
p(\mathbf{y}|\Theta) = \sum_{i=1}^{M}\pi_i\mathcal{N}(\mathbf{y};\mathbf{\mu}_i,\mathbf{\Sigma}_i),\\
%\end{equation}
%\vspace*{-20pt}
%\begin{equation}
\mathcal{N}_i(\mathbf{y}) = \dfrac{1}{\sqrt{(2\pi)^N|\mathbf{\Sigma}_i|}}e^{-\frac{(\mathbf{y}-\mathbf{\mu}_i)'{\mathbf{\Sigma}_i}^{-1}(\mathbf{y}-\mathbf{\mu}_i)}{2}}.
\end{gather}
The UBM $p(\mathbf{y}|\Theta_{\text{UBM}})$ is learned by maximising the likelihood of feature vectors extracted from the large training set using the expectation-maximisation (EM) algorithm.
%
%Learning a separate GMM for each user is suboptimal given the objective to shorten the training phase and minimize computational expenses. Furthermore, background data collection typically results in a highly unbalanced dataset in terms of variety of represented gesture patterns.
%Instead, we perform a background collection of data from hundreds of users, learn a unsupervised universal background model (UBM), which is then used as a prior for online adaptation of the model to a given user. The UBM is, in turn, formulated as a GMM and is learned by maximing likelihood of observed feature vectors from multiple users by the expecation-minimization (EM) algorithm.
%

\begin{figure}[!ttb]
\centering
%\setlength{\unitlength}{0.1\textwidth}
%\begin{picture}(10,9)
%\includegraphics[height=0.32\linewidth]{images/architecture_single.png}\hfill
%\parbox{0.4\linewidth}{\centering
%\hspace*{-20pt}
%\includegraphics[width=0.49\linewidth]{chapter05a/architecture_conv.eps}\hfill
%(a)}\hspace*{25pt}
%\parbox{0.4\linewidth}{\centering
%\includegraphics[width=0.49\linewidth]{chapter05a/architecture_dynamic.eps}
\includegraphics[width=\linewidth]{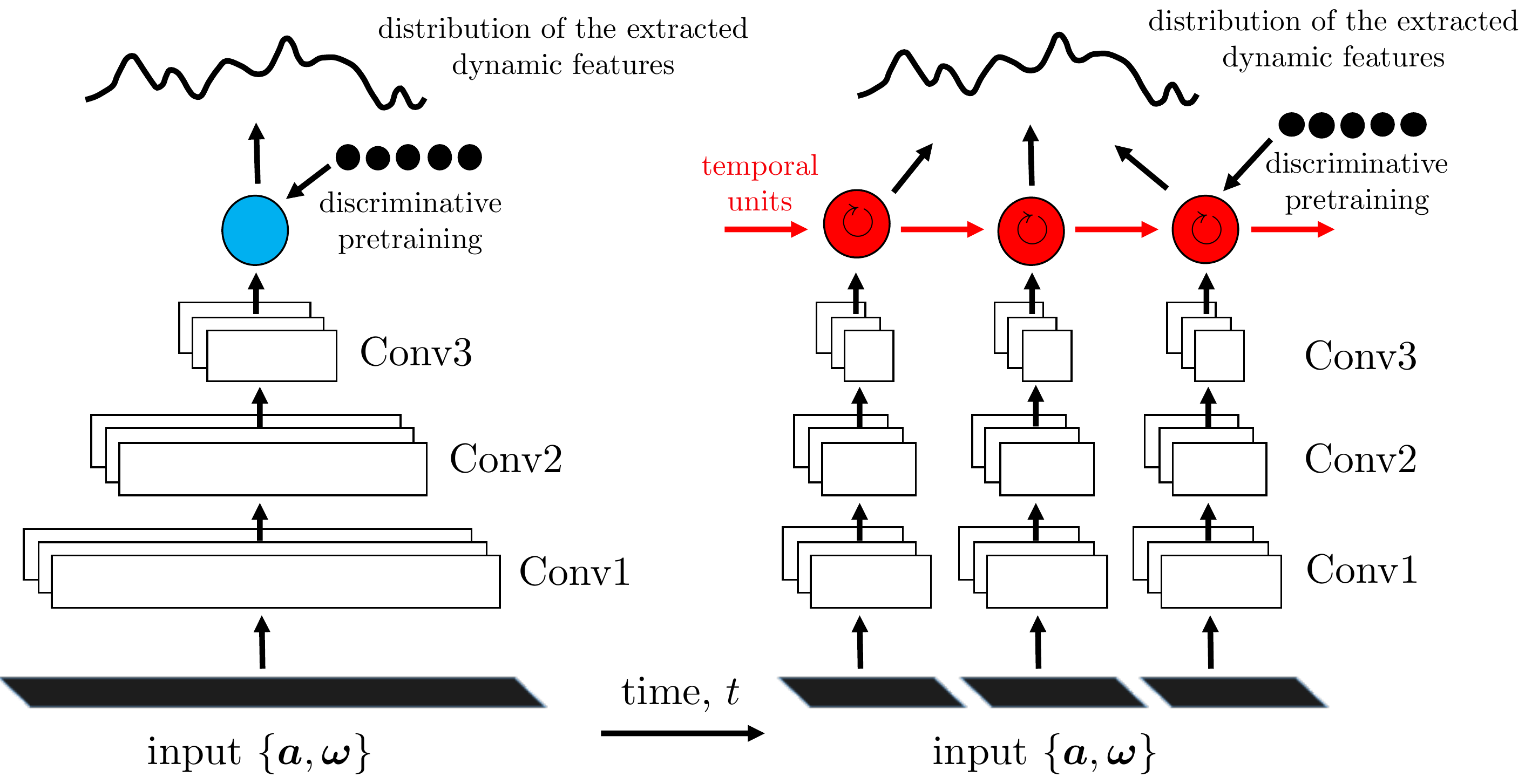}
%(b)}
\begin{tabular}{p{0.3cm}cp{3cm}cp{1cm}}
& (a) && (b)& \\
\end{tabular}
\caption[Learning data representations: convolutional learning and explicit modeling of temporal transitions.]{Learning data representations: (a) static convnet 
directly operating on sequences, aggregating temporal statistics by
temporal pooling; (b) explicitly modeling temporal transitions
with recurrent connections.}
\label{fig:architectures}%\vspace*{-10pt}
\end{figure}

The client model $p(\mathbf{y}|\Theta_{\text{client}})$ is adapted from the UBM. Both models share the same weights and covariance matrices to avoid overfitting from a limited amount of enrollment data.
Along the lines of \cite{reynolds2000}, maximum a posteriori (MAP)
adaptation of mean vectors for a given user is performed. This has an
immediate advantage over creating an independent GMM for each user,
ensuring proper alignment between the well-trained background model
and the client model by updating only a subset of parameters that are specific to the given user.
In particular,
given a set of~$Q$ enrollment samples~$\{\mathbf{y}_q\}$ from the new device, we create a
client-specific update to the mean of each mixture component~$i$ as follows:%sufficient statistics (mixture coefficient, mean and covariance matrix):\gwt{This is a bit weird, because I don't see the coefficient, mean, or covariance here (the mean is updated in the next equation but not the others)}
\begin{gather}
E_i(\{\mathbf{y}_q\})\!=\!\dfrac{1}{n_i}\sum_{q=1}^{Q}{Pr(i|\mathbf{y}_q) \mathbf{y}_q},\\
%}{n_i}\sum_{t=1}^T Pr(i|\mathbf{y}_t)\mathbf{y}_t, % \quad
%E_i(\mathbf{y}^2) = \dfrac{1}{n_i}\sum_{t=1}^T Pr(i|\mathbf{y}_t){\mathbf{y}_t}^2,
%\end{equation}
\text{where } 
n_i=\sum_{q=1}^Q\!Pr(i|\mathbf{y}_q), \;%\quad E_i(\mathbf{y}) = \dfrac{1
%\begin{equation}
Pr(i|\mathbf{y}_q)=\dfrac{\pi_ip_i(\mathbf{y}_q)}{\sum_{j=1}^M\!\pi_j p_j(\mathbf{y}_q)}.
\end{gather}
Finally, the means of all Gaussian components are updated according to the following rule: 
\begin{equation}
\hat\mu_i = \alpha_iE_i(\{\mathbf{y}_q\}) + (1-\alpha_i)\mu_i,\quad
\text{where } \alpha_i\! =\! \dfrac{n_i}{n_i + r},
\end{equation}
%$\mathbf{y}_t$ is a set of input features vector from a positive
%class (client device) and $r
\noindent where $r$ is a relevance factor balancing the background and
client models. % We set $r=4$ for all of our experiments.
In our experiments, $r$ is held fixed.

\textbf{Scoring} --- given a set of samples $Y{=}\{\mathbf{y}_s\}$ from a given device, authenticity is estimated by %the relevant question is whether the given samples have been produced by this device or from a different device. Let us note that the notion of ``device'' here should be interpreted as the user of a given device. In other words, we would like to verify whether a given device has changed hands, thus whether the given samples have been produced by a different user. 
scoring the feature vectors against the UBM and the client model, thresholding the log-likelihood ratio:
\begin{equation}
\Lambda(Y) = \log p(Y|\Theta_{\text{client}}) - \log p(Y|\Theta_{\text{UBM}}).
\label{eq:llratio}
\end{equation}
% As is standard in GMM-based speaker verification systems
As a final step, zt-score normalization \cite{ztscores} is performed to compensate for inter-session and inter-person variations and reduce the overlap between the distribution of scores from authentic users and impostors. 

%--------------------
% GWT - removing the description of zt-score normalization - move to appendix

 %--------------------

% \begin{equation}
% \Lambda_z(Y|\Theta_{\text{client}}) = \dfrac{\Lambda(Y) - \mu(Z|\Theta_{\text{client}})}{\sigma(Z|\Theta_{\text{client}})}.
% \end{equation}
%Zt-score.
%T-norm (test normalization) compensating for inter-session score variation to reduce the overlap between imposter and true score distributions.
%Z-norm (zero nomalization) compensating for inter-user score variation by normalizing them to zero mean and unit variance in order to use a single global threshold.
%$$\Lambda_z(Y|\Theta_{\text{client}}) = \dfrac{\Theta_(Y) - \mu(Z|\Theta_{\text{client}})}{\sigma(Z|\Theta_{\text{client}})},$$
%where $Y$ is a test session and $Z$ is a set of impostor sessions, both are scored against the client model $\lambda_{client}$.
%Parameters are defined once for a given user once model enrollment is completed.
%
%$$\Lambda_{zt}(Y) = \dfrac{ \Theta_z(Y) - \mu_z(Y|\Theta_t)}{\sigma_z(Y|\Theta_t)},$$
%where $\lambda_t$ is a cohort of impostor models. Parameters are estimated online for each test session.
%
%%\subsection{Total variability modelling}
%%Not yet tested.

%%% Local Variables: 
%%% mode: latex
%%% TeX-master: "nips2015"
%%% End: 

\section{Learning effective and efficient representations}
\label{sec:featurelearning}
% GWT I'd like to say effective and efficient, but it takes up two lines

%As any rigid body, a phone has 6 degrees of freedom, including translation and rotation along three perpendicular axes. For this study, the force of gravity is removed with a high-pass Butterworth filter.  encoded device signature. %But can be added and considered as a separate modality describe absolute phone orientation in space.

Learning effective and efficient data representations is key to our entire framework since its ability to perform in the real-world is defined by such criteria as latency, representational power of extracted features and inference speed of the feature extractor. The first two conditions are known to contradict each other as performance of a standalone feature typically grows with integration time \cite{moddrop}.

Two paradigms which strike a balance between representational power and speed have dominated the feature learning landscape in recent years. These are multi-scale temporal aggregation via 1-dimensional convolutional networks Fig.~\ref{fig:architectures}a, and explicit modeling of temporal dependencies via recurrent neural networks Fig.~\ref{fig:architectures}b.

The former model, popular in speech recognition \cite{hannun2014deepspeech}, involves convolutional learning of integrated temporal statistics from short and long sequences of data (referred to as ``short-term'' and ``long-term'' convnets). Short-term architectures produce outputs at relatively high rate (1~Hz in our implementation) but fail to model context. Long-term networks can learn meaningful representations at different scales, but suffer from a high degree of temporal inertia and do not generalize to sequences of arbitrary length.

Recurrent models which explicitly model temporal evolutions can generate low-latency feature vectors built in the context of previously observed user behavior. The dynamic nature of their representations allow for modeling richer temporal structure and better discrimination among users acting under different conditions. There have been a sufficiently large number of neural architectures proposed for modeling temporal dependencies in different contexts: the baseline methods compared in this  work are summarized in Fig.~\ref{fig:temporal}c. The rest of this section provides a brief description of these models. Then, Section
\ref{sec:dcwrnn} introduces a new shift-invariant model based on modified Clockwork RNNs~\cite{koutnik2014}. 

All feature extractors are first pretrained discriminatively for a
multi-device classification task, then, following removal of the
output layer the activations of the penultimate hidden layer are
provided as input (which we denote, for conciseness\footnote{This
  decision was made to avoid introducing notation to index hidden
  layers, as well as simplify and generalize the presentation in the previous
  section, where the~$\mathbf{y}$ are taken as generic temporal features.}, by~$\mathbf{y}$) to the generative model described in Section~\ref{sec:background}.
The final outputs of the background and client models are integrated over a~$30\,\textrm{sec}$ window. Accordingly, after $30\,\textrm{sec}$ the user is either authenticated or rejected.
 %and an alternative to them is learning dynamic features from short sequences combined in a continuous data stream. The two principal architectures are

\subsection{Classical RNN and Clockwork RNN}
\label{ref:rnnandcwrnn}
\noindent
The vanilla recurrent neural network (RNN) is governed by the update equation
\begin{equation} \mathbf{h}^{(t)} = \psi(\mathbf{U}\mathbf{h}^{(t-1)} + \mathbf{W}\mathbf{x}^{(t)}),
\label{eq:rnn}
\end{equation} where $\mathbf{x}$ is the input, $\mathbf{h}^{(t)}$ denotes the network's hidden state at time $t$, $\mathbf{W}$ and $\mathbf{U}$ are feed-forward and recurrent weight matrices, respectively, and $\psi$ is a nonlinear activation function, typically $\tanh$.  The output is produced combining the hidden state in a similar way, $\mathbf{o}^{(t)} = \phi(\mathbf{V}\mathbf{h}^{(t)})$, where $\mathbf{V}$ is a weight matrix.  

\begin{figure}[!t] 
\centering
%\parbox{0.5\linewidth}{
%\centering
%\begin{picture}(170,74) 
%\put(-50,-17){

%\includegraphics[width=0.64\linewidth]{rnn_unroll_short.pdf}\hspace*{-5pt}%
%\includegraphics[width=0.45\linewidth]{lstm.pdf}
\includegraphics[width=\linewidth]{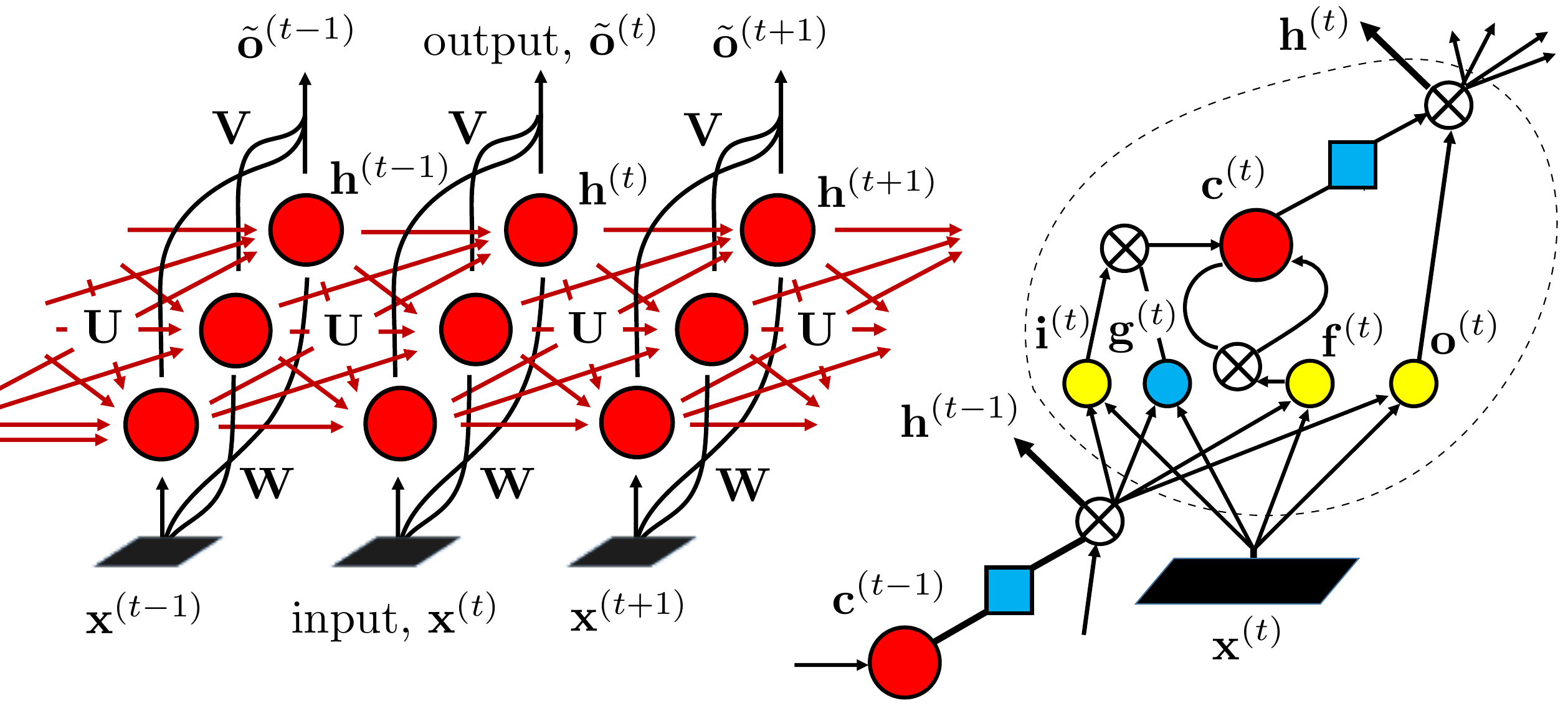}\vspace*{-5pt}
\begin{tabular}{p{1.5cm}cp{3cm}cp{3cm}}
& (a) && (b)& \\
\end{tabular}\vspace*{5pt}
\includegraphics[width=\linewidth]{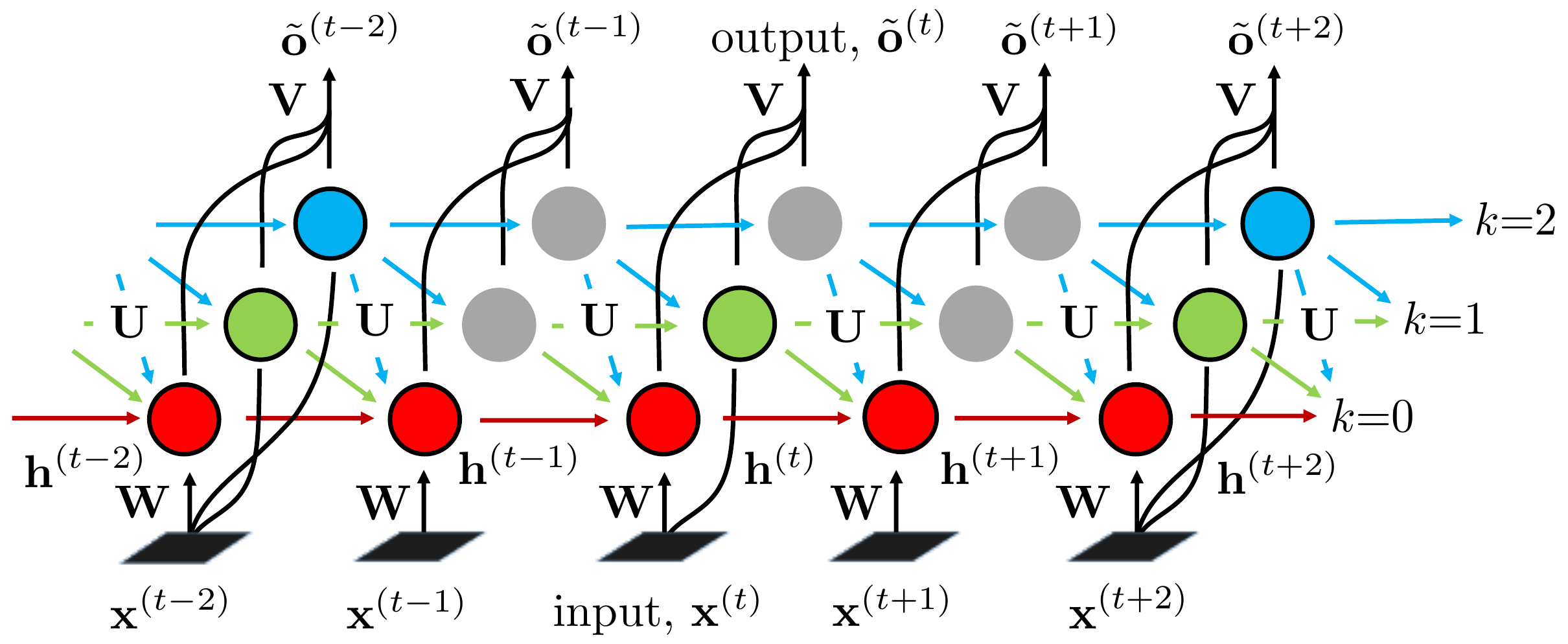}\\
\begin{tabular}{p{3.5cm}cp{6.6cm}}
& (c) & \\
\end{tabular}
%\put(42,-28){inputs} 
%\put(42,71){outputs} 
%\put(145,7){$k=0$} \put(149,19){$k=1$} \put(153,31){$k=2$}
%\end{picture}\\[35pt] 
%(d) 
%} 
%\\[5pt]
\caption{Temporal models: (a) a basic recurrent unit; (b) an LSTM unit \cite{hochreiter1997}; (c) Clockwork RNN \cite{koutnik2014} with 3 bands and a base of 2; Increasing $k$ indicates lower operating frequency. Grey color indicates inactivity of a unit.}
%\caption[Comparison of the original Clockwork RNN \cite{koutnik2014} and Dense Clockwork WRNN, proposed in this work.]{Comparison of the original Clockwork RNN, proposed by Koutnik et al. \cite{koutnik2014}, and its modification called Dense Clockwork RNN, proposed in this work.}
\label{fig:temporal}. %\vspace*{-10pt} 
\end{figure}

One of the main drawbacks of this model is that it operates at a predefined temporal scale. In the context of free motion which involves large variability in speed and changing intervals between typical gestures, this may be a serious limitation.
%A brute-force solution, an ensemble of scale-specific recurrent networks, is obviously computationally expensive.
The recently proposed Clockwork RNN (CWRNN)~\cite{koutnik2014} operates at several temporal scales which are incorporated in a single network and trained jointly. It decomposes a recurrent layer into several bands of high frequency (``fast'') and low frequency (``slow'') units (see Fig.~\ref{fig:temporal}c). Each band is updated at its own pace. The size of the step from band to band typically increases exponentially (which we call \textit{exponential update rule}) and is defined as $n^k$, where $n$ is a base and $k$ is the number of the band.  %In general, the set of step sizes can be defined arbitrarily, but due to advantages in implementation (see \cite{koutnik2014}), in the rest of the paper we will assume that all clockwork architectures have exponential bands, if not stated otherwise.

\begin{figure}[!t]
\centering
\includegraphics[width=\linewidth]{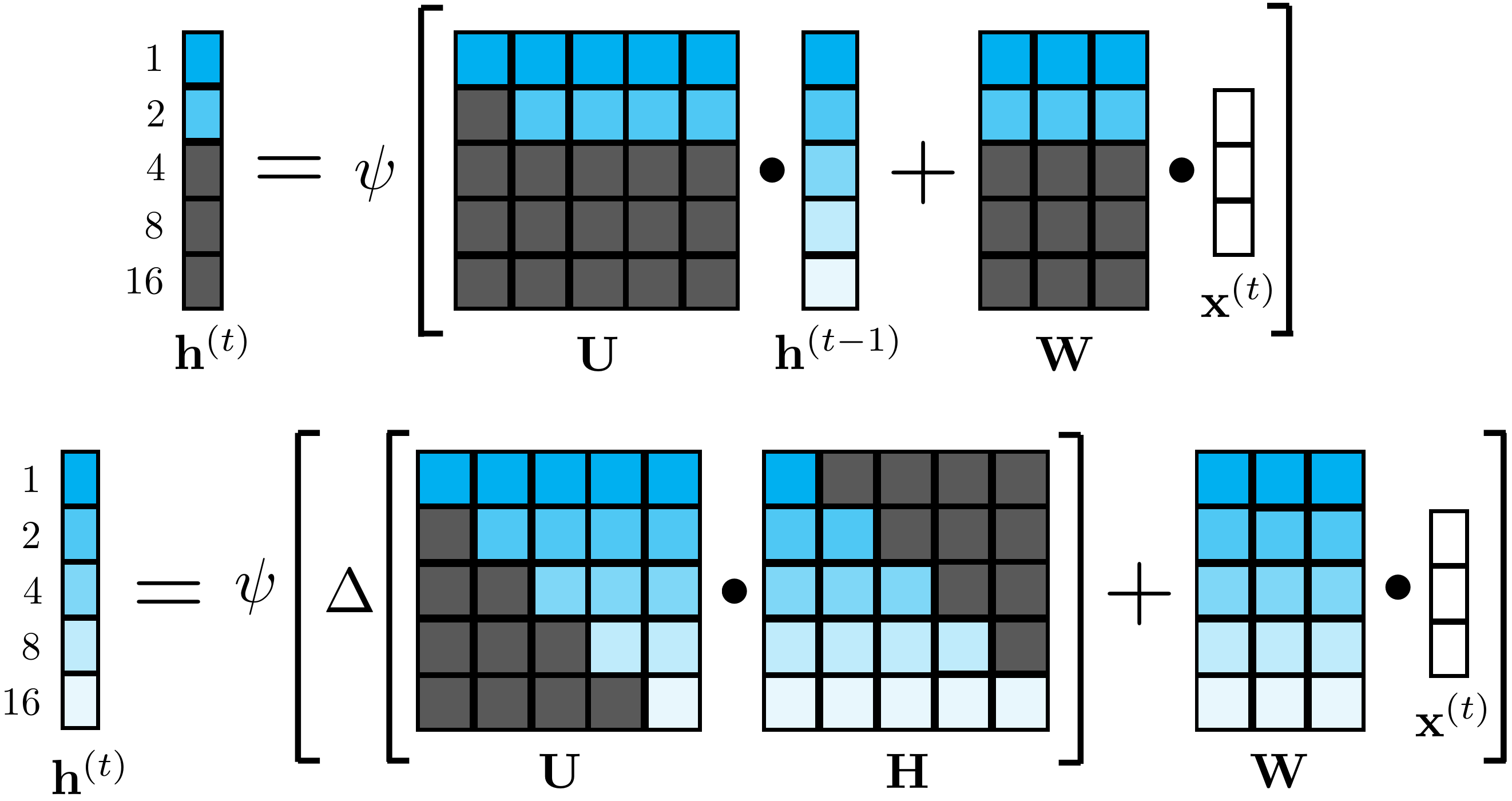}
\caption{Updates made by the Clockwork RNN \cite{koutnik2014} (top) and our proposed Dense CWRNN (bottom). Units and weights colored in blue are the ones updated or read at the example time step $t{=}6$.}
\label{fig:matrices}
\end{figure}

In the CWRNN, fast units (shown in red) are connected to all bands, benefitting from the context provided by the slow bands, while the low frequency units ignore noisy high frequency oscillations. Equation (\ref{eq:rnn}) from classical RNNs is modified, leading to a new update rule for the $k$-th band of output $\mathbf{h}$ at iteration $t$ as follows:
\begin{equation*} \mathbf{h}^{(t)}_k = \!\! \left\{
%\begin{array}{cl} \text{Equation \ref{eq:rnn}} & \text{for }\, (t\;\text{mod}\;n^k)=0,\\ h^{(t-1)}_k & \text{otherwise} .\
\begin{array}{cl} 
\!\! \psi \left(
\mathbf{U}(k) \mathbf{h}^{(t-1)}_k \! + \! \mathbf{W}(k) \mathbf{x}^{(t)}
\right) & \!\!\!\!\! \text{if }\, (t\;\text{mod}\;n^k){=}0,\\ 
\mathbf{h}^{(t-1)}_k & \!\!\!\!\! \text{otherwise} .\
\end{array} \right.
\end{equation*}
where $\mathbf{U}(k)$ and $\mathbf{W}(k)$ denote rows $k$ from matrices $\mathbf{U}$ and $\mathbf{W}$. Matrix $\mathbf{U}$ has an upper triangular structure, which corresponds to the connectivity between frequency bands. This equation is intuitively explained in the top part of Fig.~\ref{fig:matrices}, inspired from \cite{koutnik2014}. Each line corresponds to a band. At time step $t{=}6$ for instance, the first two bands $k=0$ and $k=1$ get updated. The triangular structure of the matrix results in each band getting updated from bands of lower (or equal) frequency only. In Fig.~\ref{fig:matrices}, not active rows are also shown as zero (black) in $\mathbf{U}$ and $\mathbf{W}$. In addition to multi-scale dynamics, creating sparse connections (high-to-low frequency connections are missing) reduces the number of free parameters and inference complexity.

\subsection{Long Short-Term Memory}
Long Short-Term Memory (LSTM) networks \cite{hochreiter1997}, another variant of RNNs, and their recent convolutional extensions~\cite{donahue2015,sainath2015} have proven to be, so far, the best performing models for learning long-term temporal dependencies. They handle information from the past through additional gates, which regulate how a memory cell is affected by the input signal. In particular, an input gate allows to add new memory to the cell's state, a forget gate resets the memory and an output gate regulates how gates at the next step will be affected by the current cell's state.

The basic unit is composed of input~$i$, output~$o$, forget~$f$, and input modulation~$g$ gates, and a memory cell~$c$ (see Fig.~\ref{fig:temporal}b). Each element is parameterized by corresponding feed-forward~($\mathbf{W}$) and recurrent~($\mathbf{U}$) weights and bias vectors.

%\begin{equation}
%\label{eq:u}
%\begin{gathered} %i^{(t)} = \sigma(W_Ix^{(t)} + U_Ih^{(t-1)}),\quad %f^{(t)} = \sigma(W_Fx^{(t)} + U_Fh^{(t-1)})\\ %o^{(t)} = \sigma(W_Ox^{(t)} + U_Oh^{(t-1)}),\quad %g^{(t)} = W_Gx^{(t)} + U_Gh^{(t-1)})\\ %c^{(t)} = i^{(t)} \odot c^{(t-1)} + f^{(t)}\odot g^{(t)},\quad %h^{(t)} = o^{(t)} \odot \psi(c^{(t)}),
 %\end{gathered}
 %\end{equation} % where $\odot$ denotes the Hadamard product, $x$ in input, $h$ is the output activation and $\sigma$ is a sigmoid and $\psi$ is a $tanh$ activation function. In our implementation, we do not apply additional activation to the input modulation gate.

Despite its effectiveness, the high complexity of this architecture may appear computationally wasteful in the mobile setting. Furthermore, the significance of learning long-term dependencies in the context of continuous mobile authentication is compromised by the necessity of early detection of switching between users.  Due to absence of annotated ground truth data for these events, efficient training of forgetting mechanisms would be problematic.

\subsection{Convolutional learning of RNNs}
Given the low correlation of individual frames with user identity, we
found it strongly beneficial to make the input layer convolutional
regardless of model type, thereby forcing earlier fusion of temporal
information. % This strategy is efficient for signal denoising and
% extracting characteristic motion patterns.
To simplify the presentation, we have not made convolution explicit in
the description of the methods above, however, it can be absorbed into
the input-to-hidden matrix $\mathbf{W}$.

\section{Dense convolutional clockwork RNNs}
\label{sec:dcwrnn}

Among the existing temporal models we considered, the clockwork
mechanisms appear to be the most attractive due to low computational
burden associated with them in combination with their high modeling
capacity.  However, in practice, due to inactivity of ``slow'' units
for long periods of time, they cannot respond to high frequency
changes in the input and produce outputs which, in a sense, are stale.
% , the efficiency of their training is scaled down exponentially from
% high to low frequencies. As a result, in practice
%produced by low frequency bands quickly become stale and therefore the low-frequency bands barely improve the overall network performance during test time. 
Additionally, in our setting, where the goal is to learn dynamic data representations serving as an input to a probabilistic framework, this architecture has one more weakness which stems from the fact that different bands are active at any given time step. The network will respond differently to the same input stimuli applied at different moments in time. This ``shift-variance'' convolutes the feature space by introducing a shift-associated dimension.

\begin{figure}[!t] 
\centering
%\parbox{0.5\linewidth}{
%\centering
%\begin{picture}(170,74) 
%\put(-50,-17){\includegraphics[width=70mm, trim = 120 320 120 250, clip]{dense_cwrnn.eps}}
%\put(42,-28){inputs} 
%\put(42,71){outputs} 
%\put(145,7){$k=0$} \put(149,19){$k=1$} \put(153,31){$k=2$}
%\end{picture}\\[35pt] 
%(d) 
%} 
%\\[5pt]
\includegraphics[width=\linewidth]{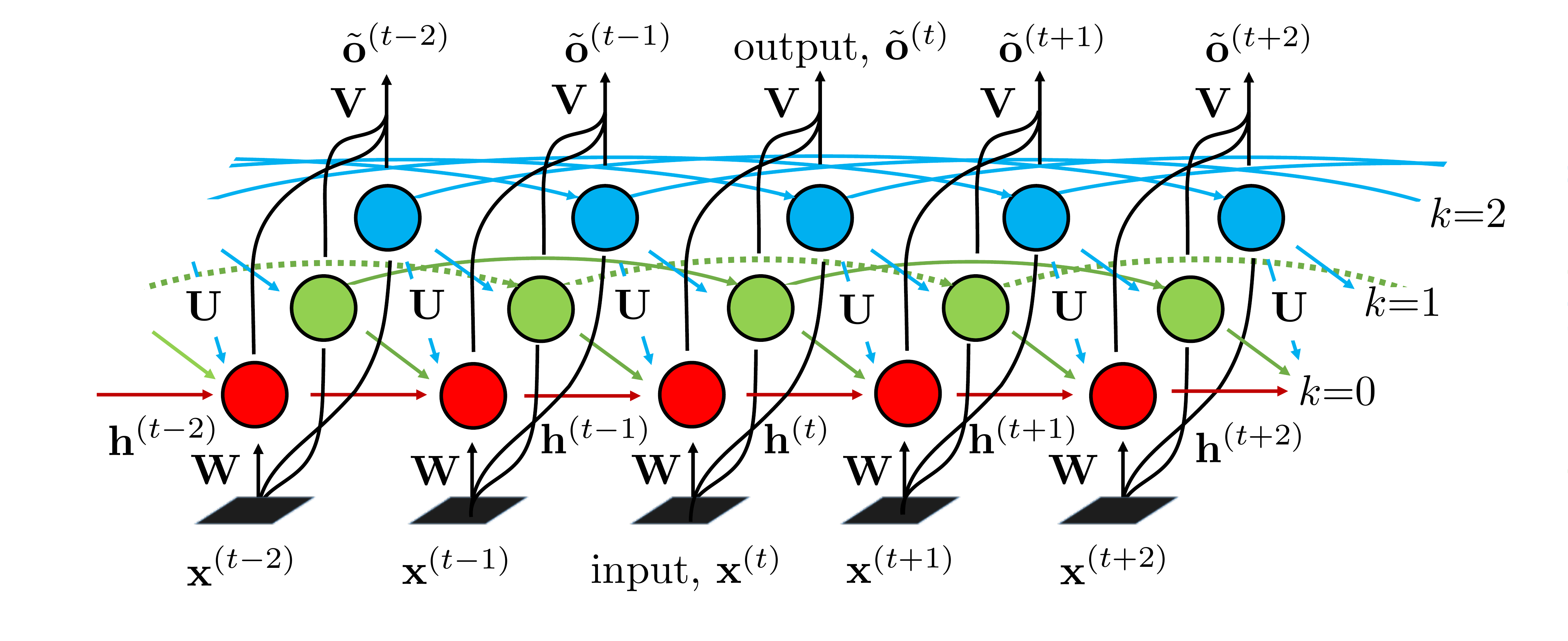}%}
\caption{Proposed dense clockwork RNN with the same parameters as the original clockwork RNN shown in Fig.~\ref{fig:temporal}a.}
\label{fig:temporaldense} %\vspace*{-10pt} 
%\vspace*{-5mm}
\end{figure}

In this work, we propose a solution to both issues, namely ``twined'' or ``dense'' clockwork mechanisms (DCWRNN, see Fig.~\ref{fig:temporaldense}), where during \textit{inference} at each scale $k$ there exist $n^k$ parallel threads shifted with respect to each other, such that at each time a unit belonging to one of the threads fires, updating its own state and providing input to the higher frequency units. 
All weights between the threads belonging to the same band are shared, keeping the overall number of parameters in the network the same as in the original clockwork architecture. Without loss of generality, and to keep the notation uncluttered of unnecessary indices, in the following we will describe a network with a single hidden unit $h_k$ per band $k$. The generalization to multiple units per band is straightforward, and the experiments were of course performed with the more general case.

The feedforward pass 
%update rule for a single hidden unit $h_{k,i}$ from band $k$ at time $t$ in this case can be expressed as follows:
%
%\begin{equation}
%h^{(t)}_{k,i} = \psi(W_k x^{(t)} + U_k h_{k,i}^{(t-n^k)})
%\end{equation}
%
for the whole dense clockwork layer (i.e.~all bands) can be given as follows:
%\begin{equation}
%h^{(t)} = \psi\left(Wx^{(t)} + {\Delta}\left[U\left(h^{(t-1)}\vert h^{(t-2)}|\ldots|h^{(t-n^k)}\right)\right]\right)
%\end{equation}
\begin{equation}
\mathbf{h}^{(t)} = \psi\left(\mathbf{W}\mathbf{x}^{(t)} + \Delta ( \mathbf{U} \mathbf{H} ) \right)
\end{equation}
where $H{=}[ \ \mathbf{h}^{(t-1)} \ldots \mathbf{h}^{(t-n^k)} \ \ldots \mathbf{h}^{(t-n^K)} \ ]$ is a matrix concatenating the history of hidden units and we define $\Delta(\cdot)$ as an operator on matrices returning its diagonal elements in a column vector. 
The intuition for this equation is given in Fig.~\ref{fig:matrices}, where we compare the update rules of the original CWRNN and the proposed DCWRNN using an example of a network with 5 hidden units each associated with one of~$K{=}5$ base~$n{=}2$ bands. To be consistent, we employ the same matrix form as in the original CWRNN paper \cite{koutnik2014}) and show components, which are inactive at time~$t$, in dark gray. 
As mentioned in Section \ref{ref:rnnandcwrnn}, in the original CWRNN, at time instant $t{=}6$, for instance, only unit $h_1$ and $h_2$ are updated, i.e.~the first two lines in Fig.~\ref{fig:matrices}. In the dense network, all hidden units $h_k$ are updated at each moment in time.

%The matrix of recurrent weights always remains in upper triangular form without additional masking. 
In addition, what was vector of previous hidden states~$\mathbf{h}^{(t-1)}$ is replaced with a lower triangular  ``history'' matrix~$\mathbf{H}$ of size~$K{\times}K$ which is obtained by concatenating several columns from the history of activations ~$\mathbf{h}$. Here, $K$ is the number of bands. Time instances are not sampled consecutively, but strided in an exponential range, i.e. $n, n^2, \ldots n^K$. Finally, the diagonal elements of the dot product of two triangular matrices form the recurrent contribution to the vector~$\mathbf{h}^{(t)}$. The feedforward contribution is calculated in the same way as in a standard RNN.

The practical implementation of the lower-triangular matrix containing the history of previous hidden activations in the DCWRNN requires usage of an additional memory buffer whose size can be given as 
$
m{=}\sum_{k=1}^K|\mathbf{h}_k|(n^{k{-}1}{-}1),
$ whereas here we have stated the general case of $|\mathbf{h}_k|{\geq}1$ hidden units belonging to band $k$.

During training, updating all bands at a constant rate is important for preventing simultaneous overfitting of high-frequency and underfitting of low-frequency bands. In practice it leads to a speedup of the training process and improved performance. Finally, due to the constant update rate of all bands in the dense network, the learned representations are invariant to local shifts in the input signal, which is crucial in unconstrained settings when the input is unsegmented. This is demonstrated in Section \ref{sec:experiments}.

%During \textit{backpropagation}, to let the gradients flow smoothly all the way back through the network, only one thread per scale updates recurrent weights of its own band, and others only provide context information to the higher frequency neurons while being treated as a constant (marked gray in Fig.~\ref{fig:temporal}).
% Feedforward weights of all bands, however, are updated at each step. 

%===========================================================
\section{Data collection}
\label{sec:database}
%\subsection{Dataset \textit{Project Abacus}}
%Continuous authentication on Nexus 5 (Android). About 1500 users. Distributed free phones, agreement, %fully informed, could delete the data for each session. Uploaded to Google Drive. One week to familiarize %theirselves with the process.

%Unobtrusive data collection. Months of natural usage, realworld noisy data. Data is recorded starting from %the moment when the phone is unlocked and until the end of a session. Sampling rate 200 Hz. To prevent %the drain of a battery, data is not recorded when the device is not moving (in this case set to 0). Two %separate thresholds on accelerometer and gyroscope streams. Synchronization on hardware timestamps. %Resampling to 50 Hz to compensate for differences in sampling rates between different devices. 

%\subsection{Project Abacus}
The dataset introduced in this work is a part of a more general multi-modal data collection effort performed by Google ATAP, known as Project Abacus. 
To facilitate the research, we worked with a third party vendor's panel to recruit and obtain consent from volunteers and provide them with LG Nexus 5 research phones which had a specialized read only memory (ROM) for data collection. Volunteers had complete control of their data throughout its collection, as well as the ability to review and delete it before sharing for research. Further, volunteers could opt out after the fact and request that all of their data be deleted. The third party vendor acted as a privacy buffer between Google ATAP and the volunteers.

The data corpus consisted of 27.62 TB of smartphone sensor signals, including images from a front-facing camera, touchscreen, GPS, bluetooth, wifi, cell antennae, etc. The motion data was acquired from three sensors: accelerometer, gyroscope and magnetometer.  This study included approximately 1,500 volunteers using the research phones as their primary devices on a daily basis. The data collection was completely passive and did not require any action from the volunteers in order to ensure that the data collected was representative of their regular usage.

Motion data was recorded from the moment after the phone was unlocked until the end of a session (i.e., until it is locked again). For this study, we set the sampling rate for the accelerometer and gyroscope sensors to 200 Hz and for the magnetometer to 5 Hz. However, to prevent the drain of a battery, the accelerometer and gyro data were not recorded when the device was at rest. This was done by defining two separate thresholds for signal magnitude in each channel. Finally, accelerometer and gyroscope streams were synchronized on hardware timestamps.

Even though the sampling rate of the accelerometer and the gyroscope was set to 200 Hz for the study, we noticed that intervals between readings coming from different devices varied slightly. To eliminate these differences and decrease power consumption, for our research we resampled all data to 50 Hz.
For the following experiments, data from 587 devices were used for
discriminative feature extraction and training of the universal background
models, 150 devices formed the validation set for tuning
hyperparameters, and another 150 devices represented ``clients'' for testing.

%%% Local Variables: 
%%% mode: latex
%%% TeX-master: "nips2015"
%%% End: 

%===========================================================
\section{Experimental results}
\label{sec:experiments}

In this section, we use an existing but relatively small inertial
dataset to demonstrate the ability of the proposed DCWRNN to learn
shift-invariant representations. We then describe our study involving
a large-scale dataset which was collected ``in the wild''.
\subsection{Visualization: HMOG dataset}

\begin{figure*}[!t]
\centering
\includegraphics[width=0.9\linewidth]{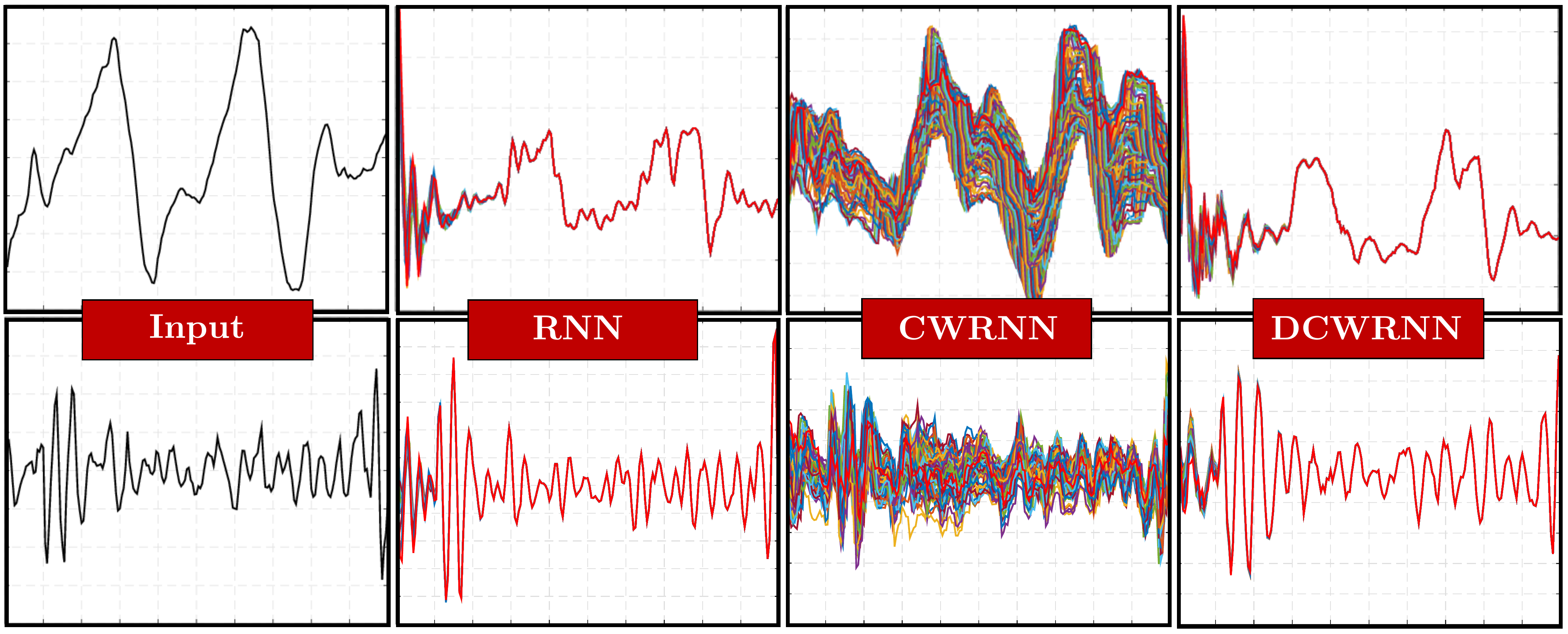}

\caption{On spatial invariance. From left to right: original sequence and traces of RNN, CWRNN and DCWRNN units. The first row: reading while walking, the second row: typing while sitting.}

\label{fig:invariance}%\vspace*{-10pt}
\end{figure*}

To explore the nature of inertial sensor signals, we performed a preliminary analysis on the HMOG dataset \cite{HMOG} containing similar data, but collected in constrained settings as a part of a lab study.
This data collection was performed with the help of 100 volunteers, each performing 24 sessions of predefined tasks, such as reading, typing and navigation, while sitting or walking. 

Unfortunately, direct application of the whole pipeline to this corpus is not so relevant due to 1) absence of task-to-task transitions in a single session and 2) insufficient data to form separate subsets for feature learning, the background model, client-specific subsets for enrollment, and still reserve a separate subset of ``impostors'' for testing that haven't been seen during training.

However, a detailed visual analysis of accelerometer and gyroscope streams has proven that the inertial data can be seen as a combination of periodic and quasi-periodic signals (from walking, typing, natural body rhythms and noise), as well non-periodic movements. This observation additionally motivates the clockwork-like architectures allowing for explicit modelling of periodic components. 

In this subsection, we describe the use of HMOG data to explore the shift-invariance of temporal models that do not have explicit reset gates (i.e.~RNN, CWRNN and DCWRNN). For our experiment, we randomly selected 200 sequences of normalized accelerometer magnitudes and applied three different networks each having 8 hidden units and a single output neuron. All weights of all networks were initialized randomly from a normal distribution with a fixed seed. For both clockwork architectures we used a base 2 exponential setting rule and 8 bands. 

Finally, for each network we performed 128 runs (i.e.~$2^{K-1}$) on a shifted input: for each run $x$ the beginning of the sequence was padded with $x{-}1$ zeros. The resulting hidden activations were then shifted back to the initial position and superimposed. Fig.~\ref{fig:invariance} visualizes the hidden unit traces for two sample sequences from the HMOG dataset, corresponding to two different activities: reading while walking and writing while sitting. The figure shows that the RNN and the dense version of the clockwork network can be considered shift-invariant (all curves overlap almost everywhere, except for minor perturbance at the beginning of the sequence and around narrow peaks), while output of the 
CWRNN is highly shift-dependent.

For this reason, in spite of their atractiveness in the context of multi-scale periodic and non-periodic signals, the usage of the CWRNN for the purpose of feature learning from unsegmented data may be suboptimal due to high shift-associated distortion of learned distributions, which is not the case for DCWRNNs.

\subsection{Large-scale study: Google Abacus dataset}
We now evaluate our proposed authentication framework on the real-world dataset described in Section \ref{sec:database}.
Table~\ref{table:networkparameters} 
in the Appendix provides architectural hyper-parameters chosen for two 1-d Convnets (abbreviated ST and LT for short and log-term) as well as the recurrent models. To make a fair comparison, we set the number of parameters to be approximately the same for all of the RNNs.
%
%All networks have 3 input convolutional layers with 25 1D temporal filters, followed by max pooling. Filters are of size $9\times 1$ for the static convnets and $7\times 1$ for temporal models. The subsequent two layers are fully connected with respective sizes of 2000 and 1000 units. The two fully connected layers in ST and LT Convnets are replaced with a single recurrent layer in the temporal architectures. 
%
The ST Convnet is trained on sequences of 50 samples (corresponding to a \SI{1}{s} data stream), LT Convnets take as input 500 samples (i.e.~\SI{10}{s}). All RNN architectures are trained on sequences of 20 blocks of 50 samples each with 50\% of inter-block overlap to ensure smooth transitions between blocks (therefore, also a \SI{10}{s} duration). For the dense and sparse clockwork architectures we set the number of bands to 3 with a base of 2.
All layers in all architectures use $\tanh$ activations.
During training, the networks produce a softmax output per block in
the sequence, rather than only for the last one. The
mean per-block negative log likelihood loss taken over all blocks is minimized.

\setlength{\tabcolsep}{4pt}
\begin{table}[!t] 
\caption{\label{table:background_models} Performance and model complexity of the feature extractors. These results assume one user per device and accuracy is defined based on whether or not the user is in the top 5\% of classes according to the output distribution.}
\centering
\begin{tabular}{l|c|c}
\toprule
\noalign{\smallskip}
\multicolumn{3}{c}{\textbf{Feature learning: evaluation}}\vspace*{3pt}\\
\hline\noalign{\smallskip}
\hfill Model \hfill & \; Accuracy, \% \;& \;\;\# parameters\;\; \\
\noalign{\smallskip}
\hline
\noalign{\smallskip}
ST Convnet 	& $37.13$	&$6\,102\,137$\\% & test\\
LT Convnet	& $56.46$	&$6\,102\,137$\\% & test\\
%Conv-RNN  	& $64.59$	&$3\,289\,187$\\% & test\\
%Conv-CWRNN 	& $68.83$&$3\,019\,187$\\% & test\\
%\textbf{Conv-DCWRNN}\! & $\mathbf{69.41}$	&$\mathbf{3\,019\,187}$\\% & test\\
%\textbf{Conv-LSTM} 	& $\mathbf{71.96}$	&$\mathbf{11\,392\,187}$\\% & test\\
Conv-RNN  	& $64.57$	&$1\,960\,295$\\% & test\\
Conv-CWRNN 	& $68.83$&$1\,964\,254$\\% & test\\
Conv-LSTM 	& $68.92$	&$1\,965\,403$\\% & test\\
\textbf{Conv-DCWRNN}\;\; & $\boldsymbol{69.41}$	&$\boldsymbol{1\,964\,254}$\\% & test\\
\bottomrule
\end{tabular}
% obtained for each session from a test subset of the training data.
\end{table}
\setlength{\tabcolsep}{1.4pt}

\setlength{\tabcolsep}{4pt}
\begin{table}[!t] \centering
\caption{\label{table:googlepredictions} 
Performance of the GMM-based biometric model using different types of deep neural architectures. EER is given on the validation set, while HTER is estimated on the final test set using the same threshold.}
\begin{tabular}{l|c|c}
\toprule
\noalign{\smallskip}
\multicolumn{3}{c}{\textbf{User authentication: evaluation}}\vspace*{3pt}\\
\hline\noalign{\smallskip}
\hspace*{\fill} Model \hspace*{\fill} & \;\;EER, \%\;\; &\;\;HTER, \%\;\; \\% & Subset\\
\noalign{\smallskip}
\hline
\noalign{\smallskip}
%Feature extraction + TV      	& $0.XXX$	& $0.XXX$ \\% & test\\
Raw features 	& $36.21$	& $42.17$ \\% & test\\
%Conv. learning + TV      	& $0.XXX$	& $0.XXX$ \\% & test\\
ST Convnet   	& $32.44$	& $34.89$ \\% & test\\
LT Convnet 	& $28.15$	& $29.01$ \\% & test\\
%Seq. learning + TV      	& $0.XXX$	& $0.XXX$ \\% & test\\
%Conv-LSTM	& $22.92$	& $22.73$ \\% & test\\
%\textbf{Conv-LSTM, zt-norm} 	& $\mathbf{20.11}$	& $\mathbf{20.29}$ \\% & test\\
%\textit{Conv-LSTM (per device)}	& $\mathit{16.29}$	& $\mathit{16.42}$ \\% & test\\
%\textit{Conv-LSTM (per session)} 	& $\mathit{9.08}$	& $\mathit{9.56}$ \\% & test\\
Conv-RNN	& $22.32$	& $22.49$ \\% & test\\
Conv-CWRNN	& $21.52$	& $21.92$ \\% & test\\
Conv-LSTM	& $21.13$	& $21.41$ \\% & test\\
Conv-DCWRNN	& $20.01$	& $20.52$ \\% & test\\
\textbf{Conv-DCWRNN, zt-norm} 	& $\boldsymbol{18.17}$	& $\boldsymbol{19.29}$ \\% & test\\
\textit{Conv-DCWRNN (per device)}	& ${15.84}$	& ${16.13}$ \\% & test\\
\textit{Conv-DCWRNN (per session)}\;\; 	& ${8.82}$	& ${9.37}$ \\% & test\\
\bottomrule
\end{tabular}
\end{table}
\setlength{\tabcolsep}{1.4pt}

The dimensionality of the feature space produced by each of the networks is PCA-reduced to 100.
GMMs with 256 mixture components are trained for 100 iterations after initialization with k-means (100 iterations). MAP adaptation for each device is performed in 5 iterations with a relevance factor of 4 (set empirically). For zt-score normalization, we exploit data from the same training set and create 200 t-models and 200 z-sequences from non-overlapping subsets. Each t-model is trained based on UBM and MAP adaptation. All hyper-parameters were optimized on the validation set.

The networks are trained using stochastic gradient descent, dropout in fully connected layers, and negative log likelihood loss. In the temporal architectures we add a mean pooling layer before applying the softmax. Each element of the input is normalized to zero mean and unit variance.
All deep nets were implemented with Theano \cite{theano} and trained on 8 Nvidia Tesla K80 GPUs. UBM-GMMs were trained with the Bob toolbox \cite{bob2012} and did not employ GPUs.

%- Smartphone ?
%Deep learning on mobile devices \cite{lane2015}
%Caffe \cite{jia2014caffe}

\textbf{Feature extraction} --- we first performed a quantitative evaluation of the effectiveness of feature extractors alone as a multi-class classification problem, where one class corresponds to one of 587 devices from the training set. This way, one class is meant to correspond to one ``user'', which is equal to ``device'' in the training data (assuming devices do not change hands). To justify this assumption, we manually annotated periods of non-authentic usage based on input from the smartphone camera and excluded those sessions from the test and training sets. Experiments showed that the percentage of such sessions is insignificant and their presence in the training data has almost no effect on the classification performance.

Note that for this test, the generative model was not considered and the feature extractor was simply evaluated in terms of classification accuracy. To define accuracy, we must consider that  human kinematics sensed by a mobile device can be considered as a weak biometric and used to perform a soft clustering of users in behavioral groups. To evaluate the quality of each feature extractor in the classification scenario, for each session we obtained aggregated probabilities over target classes and selected the 5\% of classes with highest probability (in the case of 587 classes, the top 5\% corresponds to the 29 classes). After that, the user behavior was considered to be interpreted correctly if the ground truth label was among them. 

The accuracy obtained with each type of deep network with its corresponding number of parameters is reported in Table \ref{table:background_models}. These results show that the feed forward convolutional architectures generally perform poorly, while among the temporal models the proposed dense clockwork mechanism Conv-DCWRNN appeared to be the most effective, while the original clockwork network (Conv-CWRNN) was slightly outperformed by the LSTM.
 
\textbf{Authentication evaluation} ---
when moving to the binary authentication problem, an optimal balance of false rejection and false acceptance rates, which is not captured by classification accuracy, becomes particularly important. We use a validation subset to optimize the generative model for the minimal equal error rate (EER). The obtained threshold value $\theta_{EER}$ is then used to evaluate performance on the test set using the half total error rate (HTER) as a criterion:
$
\text{HTER} = 1/2 [ \text{FAR}(\theta_{EER})+\text{FRR}(\theta_{EER}) ],
$
where FAR and FRR are false acceptance and false rejection rates, respectively. For the validation set, we also provide an average of per-device and per-session EERs (obtained by optimizing the threshold for each device/session separately) to indicate the upper bound of performance in the case of perfect score normalization (see italicized rows in Table \ref{table:googlepredictions}). 

An EER of $20\%$ means that $80\%$ of the time the correct user is using the device, s/he is authenticated, only by the way s/he moves and holds the phone, not necessarily interacting with it. It also means that $80\%$ of the time the system identifies the user, it was the correct one. These results align well with the estimated quality of feature extraction in each case and show that the context-aware features can be efficiently incorporated in a generative setting.

To compare the GMM performance with a traditional approach of retraining, or finetuning a separate deep model for each device (even if not applicable in a mobile setting), we randomly drew 10 devices from the validation set and replaced the output layer of the pretrained LSTM feature extractor with a binary logistic regression. The average performance on this small subset was $ 2\%$ inferior with respect to the GMM, due to overfitting of the enrollment data and poor generalization to unobserved activities. This is efficiently handled by mean-only MAP adaptation of a general distribution in the probabilistic setting.

Another natural question is whether the proposed model learns something specific to the user ``style'' of performing tasks rather than a typical sequence of tasks itself. To explore this, we performed additional tests by extracting parts of each session where all users interacted with the same application (a popular mail client, a messenger and a social network application).
We observed that the results were almost identical to the ones previously obtained on the whole dataset, indicating low correlation with a particular activity.

%\nn{[On pure NN approaches]
%As we mentioned, straightforward training (or fine tuning) of one network per device is not feasible, as training has to be done on a smartphone. Also, this strategy gave worse results in our experiments (tested on 10 random devices). An alternative could be a conditional neural model taking into account a background prior and user-specific information (either by biasing hidden units or modulating weights). This would be interesting future research.}

\section{Model adaptation for a visual context}

\setlength{\tabcolsep}{4pt}
\begin{table}[t] \centering
\centering
\caption{\label{table:googlevisualexperiments} Performance of the proposed DCWRNN architecture on the \textit{Chalearn 2014 Looking at People dataset} (mocap modality). Network parameters: input $183{\times}9$, conv. layer~$25{\times}3{\times}1$, 2 fully connected layers with $700$~units, the recurrent layer (RNN-280, CWRNN-300, DCWRNN-300, Small LSTM-88, Large LSTM-300), $21$~output class. } 
\begin{tabular}{l|c|c|c}
\toprule
 \noalign{\smallskip}
\multicolumn{4}{c}{\textbf{ChaLearn 2014: sequential learning}}\vspace*{3pt}\\
 \hline\noalign{\smallskip}
\hfill Model \hfill & \;\;Jaccard Index\;\; & \;\;Accuracy\;\; & \;\;N parameters\;\; \\
\noalign{\smallskip}
\hline
\noalign{\smallskip}
Single network \cite{moddrop}	& $0.827$	& $91.62$  & $1\,384\,621$\\
Ensemble \cite{moddrop}		& $0.831$	& $91.96$  & $4\,153\,863$\\[2pt]
\hline
\noalign{\smallskip}
Conv-RNN  					& $0.826$         & $91.79$  & $3\,974\,581$\\
Small Conv-LSTM       			& $0.815$         & $91.50$  & $3\,976\,863$\\
Large Conv-LSTM       			& $0.825$	    & $91.89$  & $4\,900\,621$\\
Conv-CWRNN 					& $0.834$         & $92.38$  & $3\,972\,496$\\
\textbf{Conv-DCWRNN}\! 			& $\boldsymbol{0.841}$	  & $\boldsymbol{93.02}$  & $3\,972\,496$\\
\hline
\end{tabular}\vspace*{5pt}
\end{table}

Finally, we would like to stress that the proposed DCWRNN framework can also be applied to other sequence modeling tasks, including the visual context.
%One of the main concerns in the reviews is that the proposed model was applied to a non-visual task. However, as R2 pointed out, \emph{``the models and methods are familiar to the vision community''}, i.e.~the models discussed 
The described model is not specific to the data type and there is no particular reason why it cannot be applied to the general human kinematic problem (such as, for example, action or gesture recognition from motion capture).

To support this claim, we have conducted additional tests of the proposed method within a task of \textit{visual gesture recognition}.
Namely, we provide results on the \textit{motion capture (mocap)} modality of the \textit{ChaLearn 2014 Looking at People} gesture dataset \cite{escalera2014chalearn}. This dataset contains about 14000 instances of Italian conversational gestures with the aim to detect, recognize and localize gestures in continuous noisy recordings. Generally, this corpus comprises multimodal data captured with the Kinect and therefore includes RGB video, depth stream and mocap data. However, only the last channel is used in this round of experiments. 
The model evaluation is performed using the Jaccard index, penalizing for errors in classification as well as imprecise localization. %To the best of our knowledge, this is the only gesture dataset containing enough data for deep learning.

Direct application of the GMM to gesture recognition is suboptimal (as the vocabulary is rather small and defined in advance), therefore, in this task we perform end-to-end discriminative training of each model to evaluate the effectiveness of \textit{feature extraction} with the Dense CWRNN model.

In the spirit of~\cite{moddrop} (the method ranked first\textsuperscript{st} in the ECCV 2014 ChaLearn competition), we use the same skeleton descriptor as input. However, as in the described authentication framework, the input is fed into a convolutional temporal architecture instead of directly concatenating frames in a spatio-temporal volume. 
The final aggregation and localization step correspond to \cite{moddrop}. 
Table~\ref{table:googlevisualexperiments} reports both the Jaccard index and per-sequence classification accuracy and shows that in this application, the proposed DCWRNN also outperforms the alternative solutions.

\section{Conclusion}
From a modeling perspective, this work has demonstrated that temporal architectures are particularly efficient for learning of dynamic features from a large corpus of noisy temporal signals, and that the learned representations can be further incorporated in a generative setting. With respect to the particular application, we have confirmed that natural human kinematics convey necessary information about person identity and therefore can be useful for user authentication on mobile devices. The obtained results look particularly promising, given the fact that the system is completely non-intrusive and non-cooperative, i.e.~does not require any effort from the user's side.%, even including actual interaction with the phone.

Non-standard weak biometrics are particularly interesting for providing the context in, for example, face recognition or speaker verification scenarios. Further augmentation with data extracted from keystroke and touch patterns, user location, connectivity and application statistics (ongoing work) may be a key to creating the first secure non-obtrusive mobile authentication framework.

Finally, in the additional round of experiments, we have demonstrated that the proposed Dense Clockwork RNN can be successfully applied to other tasks based on analysis of sequential data, such as gesture recognition from visual input.

% or
\appendix
%\addtocounter{table}{1} % increment table counter to be consistent
                        % with paper numbering

\vspace*{-20pt}
 \setlength{\tabcolsep}{4pt}
\begin{table*}[!tbh] 
\centering
\caption{\label{table:networkparameters} Hyper-parameters: values in
 parentheses are for short-term (ST) Convnets when different from
 long-term (LT).}
 \begin{tabular}{c|c|c||c|c}
 \hline\noalign{\smallskip}
 Layer & Filter size / \# of units & Pooling & Filter size / \# of units & Pooling\\
 \hline
 \noalign{\smallskip}&
 \multicolumn{2}{c||}{Convolutional feature learning} &
 \multicolumn{2}{c}{Sequential feature learning}\\
 \hline
 \noalign{\smallskip}
 Input & $500{\times}14$ ($50\times14$) 	       & - 			& $10{\times}50{\times}14$ &   -\\
 Conv1 & $25{\times}9{\times}1$ & $8{\times}1$  ($2{\times}1$)  &  $25{\times}7{\times}1$ & $2{\times}1$\\
 Conv2 & $25{\times}9{\times}1$ & $4{\times}1$ ($1{\times}1$)   & $25{\times}7{\times}1$ & $2{\times}1$\\
 Conv3 & $25{\times}9{\times}1$ & $1{\times}1$   & $25{\times}7{\times}1$ & $1{\times}1$\\
 FCL1     & $2000$                         &  -                  & -                         &  - \\
 FCL2     & $1000$                         &- 		& -                      & - \\
 Recurrent    & -                       &- 		& RNN 894, LSTM 394,                         & - \\
   &                       & 		& CWRNN and DCWRNN 1000                        & - \\
 %Output & $587$                           &-               & $587$                            & - \\
  \hline
 \end{tabular}

 %\end{table*}
 \setlength{\tabcolsep}{1.4pt}
\end{table*}\vspace*{20pt}

In this section, we provide additional detail for reproducability that
was not provided in the main text.

\subsection*{Hyper-parameter selection}

Table \ref{table:networkparameters} provides the complete set of
hyper-parameters that were chosen based on a held-out validation
set. For convolutional nets, we distinguish between convolutional
layers (Conv, which include pooling) and fully-connected layers
(FCL). For recurrent models, we report the total number of units (in
the case of CWRNN and DCWRNN, over all bands).

\subsection*{Details on zt-normalization}

Here we provide details on the zt-normalization that were not given in
Section 3. %\ref{sec:background}. 
% There are outstanding issues
% The subscript t is overloaded - it means time
% The first equation is ok, but the second equation is ambiguous, i.e. what is \mu_z, what is \sigma_z, etc.?
Recall that we estimate authenticity, given a set of motion features
by scoring the features against a universal background model (UBM) and
client model. Specifically, we threshold the log-likelihood ratio in
Eq. 7. %~\ref{eq:llratio}. 
An offline z-step (zero normalization) compensates for inter-model variation by normalizing the scores produced by each client model to have zero mean and unit variance in order to use a single global threshold:
\begin{equation}
\Lambda_z(Y|\Theta_{\text{client}}) = \dfrac{\Lambda(Y) - \mu(Z|\Theta_{\text{client}})}{\sigma(Z|\Theta_{\text{client}})},
\end{equation}
\noindent where $Y$ is a test session and $Z$ is a set of impostor sessions. Parameters are defined for a given user once model enrollment is completed.
Then, the $\mathcal{T}$-norm (test normalization) compensates for inter-session differences by scoring a session against a set of background $\mathcal{T}$-models.
\begin{equation}
\Lambda_{zt}(Y) = \dfrac{ \Lambda_z(Y|\Theta_{\text{client}}) - \mu_z(Y|\Theta_{\mathcal{T}})}{\sigma_z(Y|\Theta_{\mathcal{T}})}.
\end{equation}
The T-models are typically obtained through MAP-adaptation from the universal background model in the same way as all client models, but using different subsets of the training corpus. The Z-sequences are taken from a part of the training data which is not used by the T-models.

%%% Local Variables:
%%% mode: latex
%%% TeX-master: "main"
%%% End:

\section*{Acknowledgment}
The authors would like to thank their colleagues Elie Khoury, Laurent El Shafey, Sébastien Marcel and Anupam Das for valuable discussions.


\ifCLASSOPTIONcaptionsoff
  \newpage
\fi

\bibliographystyle{ieee}
\bibliography{biometrics}

\begin{IEEEbiography}[{\includegraphics[width=1in,height=1.25in,clip,keepaspectratio]{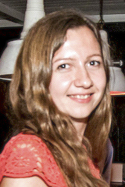}}]{Natalia Neverova}
is a PhD candidate at INSA de Lyon and LIRIS (CNRS, France) working in the area of gesture and action recognition with emphasis on multi-modal aspects and deep learning methods. She is advised by Christian Wolf  and Graham Taylor, and her research is a part of Interabot project in partnership with Awabot SAS. She was a visiting researcher at University of Guelph in 2014 and at Google in 2015. 
%She holds a Europeen CIMET Erasmus Mundus MSc degree with excellent distinction.\\
\end{IEEEbiography}
\vspace*{-10mm}
\begin{IEEEbiography}[{\includegraphics[width=1in,height=1.25in,clip,keepaspectratio]{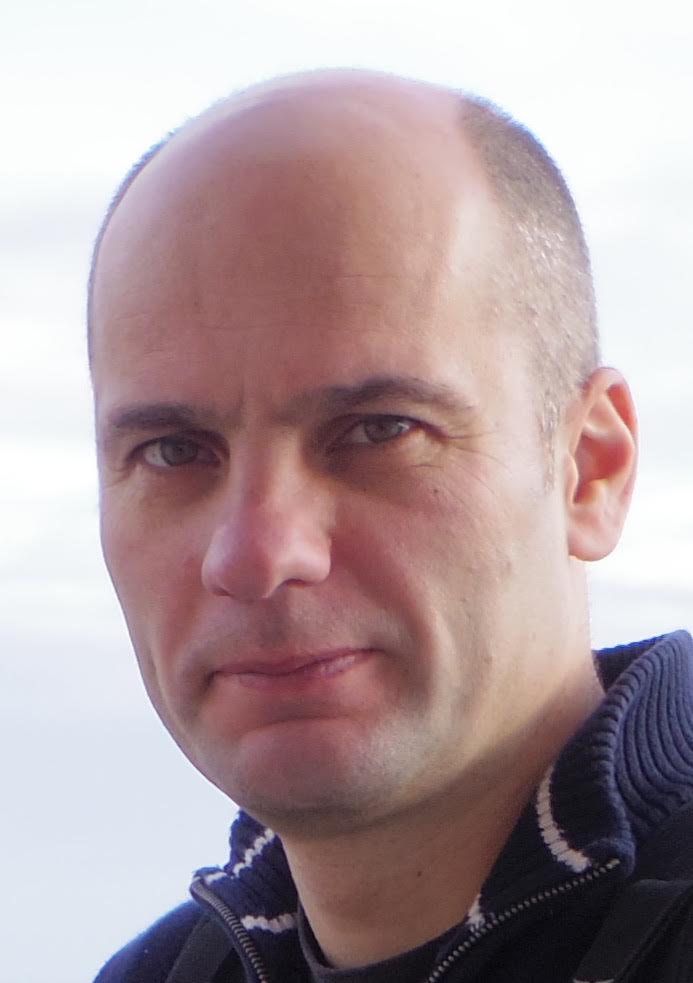}}]{Christian Wolf}
is assistant professor at INSA de Lyon and LIRIS, CNRS, since 2005. He is interested in computer vision and machine learning, especially the visual analysis of complex scenes in motion, 
 structured models, graphical models and deep learning. 
He received his MSc in computer science from Vienna University of Technology in 2000, and a PhD from the National Institute of Applied Science (INSA de Lyon), in 2003. In 2012 he obtained the habilitation diploma.
\end{IEEEbiography}
\vspace*{-11mm}
\begin{IEEEbiography}[{\includegraphics[width=1.2in,height=1.25in,clip,keepaspectratio]{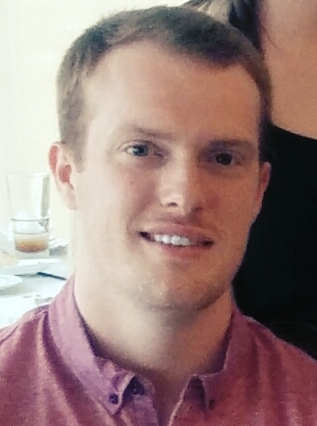}}]
{Griffin Lacey} is a Masters of Applied Science student at the University of Guelph.  His primary research interests are in developing tools and techniques to support deep learning on FPGAs.  Recently, he acted as a visiting researcher at Google.  Griffin holds a BENG in Engineering Systems and Computing from the University of Guelph.
\end{IEEEbiography}
\vspace*{-8mm}
\begin{IEEEbiography}[{\includegraphics[width=1.2in,height=1.25in,clip,keepaspectratio]{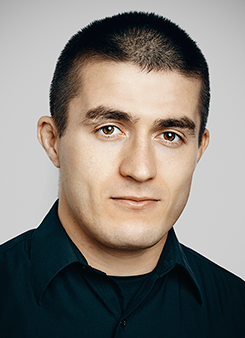}}]
{Lex Fridman} is a postdoctoral associate at the Massachusetts Institute of Technology (MIT). He received his BS, MS, and PhD from Drexel University. His research interests include machine learning, decision fusion, and computer vision applied especially to detection and analysis of human behavior in semi-autonomous vehicles.
\end{IEEEbiography}
\begin{IEEEbiography}[{\includegraphics[width=1.2in,height=1.25in,clip,keepaspectratio]{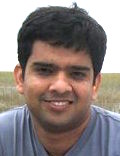}}]
{Deepak Chandra}
heads authentication at the Machine Intelligence and Research group at Google. The project aims at completely redefining authentication for digital and physical world.  Prior to this he was the program lead in Google’s Advanced Technology and Projects (ATAP) organization, where he heads all product, engineering, and design for mobile authentication projects. Deepak defined company wide authentication strategy for Motorola prior to leading the efforts at Google. He has developed multiple wearable authentication products including Motorola Skip and Digital Tattoo. 
\end{IEEEbiography}
\vspace*{-10mm}
\begin{IEEEbiography}[{\includegraphics[width=1in,height=1.25in,clip,keepaspectratio]{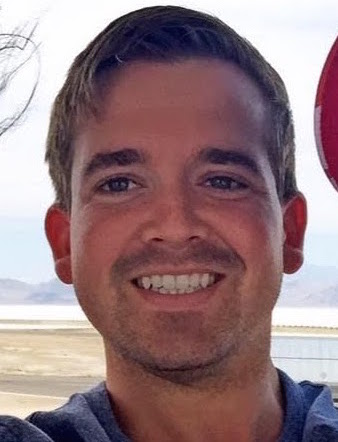}}]
{Brandon Barbello}
is a Product Manager at Google Research \& Machine Intelligence, where he works on privacy-sensitive on-device machine learning. He was previously at Google Advanced Technology and Projects (ATAP) on the Project Abacus team, where he managed efforts to develop a multimodal continuous authentication system for smartphones. Prior to Google, Brandon co-founded four companies across electronics, fintech, and private equity.
\end{IEEEbiography}
\vspace*{-10mm}

\begin{IEEEbiography}[{\includegraphics[width=1in,height=1.25in,clip,keepaspectratio]{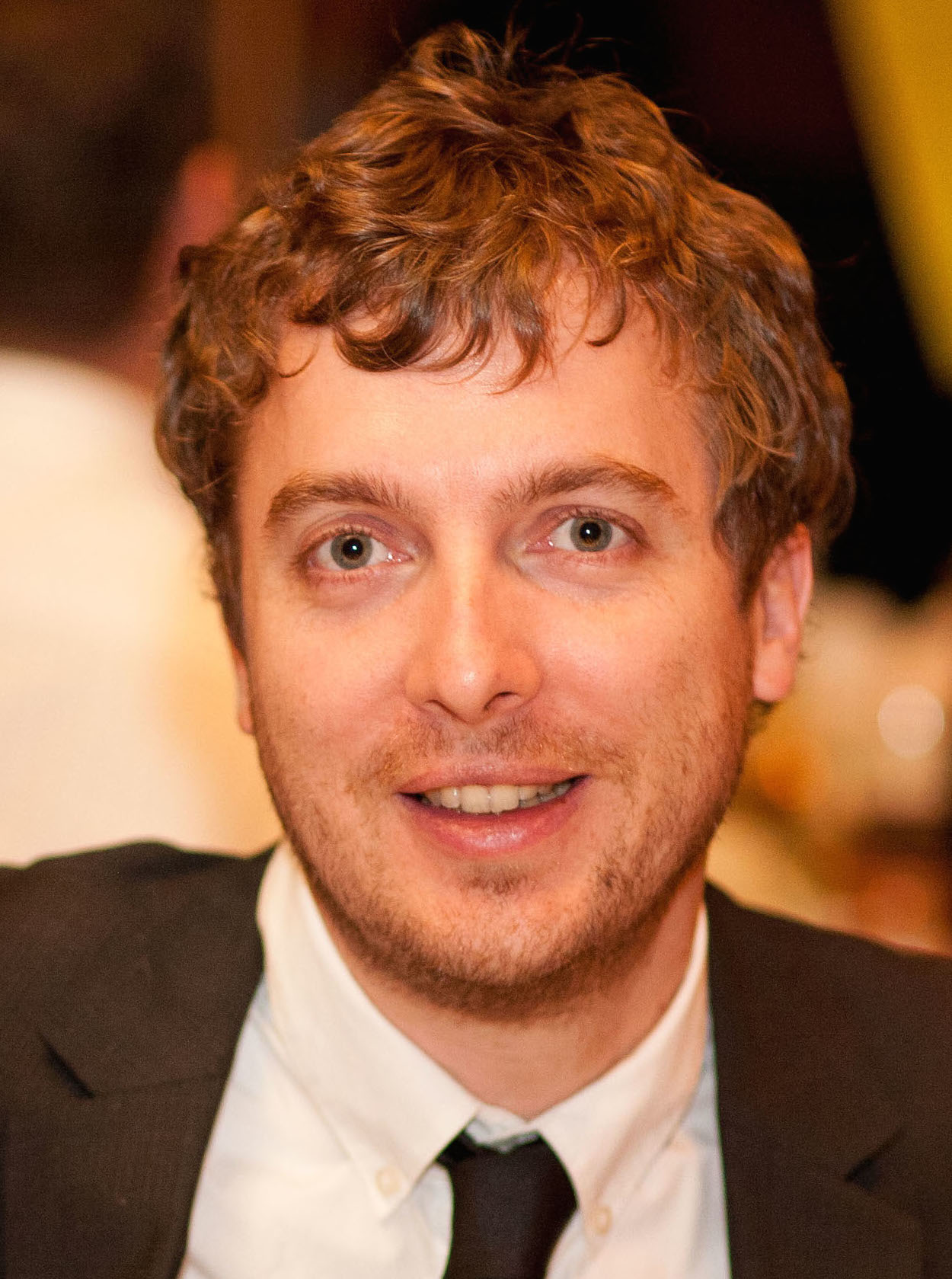}}]{Graham Taylor}
is an assistant professor at University of Guelph. He is interested in statistical machine learning and biologically-inspired computer vision, with an emphasis on unsupervised learning and time series analysis. He completed his PhD at the University of Toronto in 2009, where his thesis co-advisors were Geoffrey Hinton and Sam Roweis. He did a postdoc at NYU with Chris Bregler, Rob Fergus, and Yann LeCun.
\end{IEEEbiography}
\vspace*{120mm}\phantom{t}\newpage

%

%\begin{IEEEbiography}{Natalia Neverova}
%\end{IEEEbiography}

%\begin{IEEEbiography}{Christian Wolf}
%\end{IEEEbiography}

%\begin{IEEEbiography}{Griffin Lacey}
%\end{IEEEbiography}

%\begin{IEEEbiography}{Lex Fridman}
%\end{IEEEbiography}

%\begin{IEEEbiography}{Deepak Chandra}
%\end{IEEEbiography}

%\begin{IEEEbiography}{Brandon Barbello}
%\end{IEEEbiography}

%\begin{IEEEbiography}{Graham Taylor}
%\end{IEEEbiography}


%\vfill


\end{document}